\documentclass[10pt,twocolumn,letterpaper]{article}

\usepackage{iccv}
\usepackage{times}
\usepackage{epsfig}
\usepackage{graphicx}
\usepackage{amsmath}
\usepackage{amssymb}
\usepackage{subfigure}
\usepackage[ruled,vlined]{algorithm2e}
\usepackage{booktabs}
\usepackage{multirow}
\usepackage{pifont}
\usepackage{soul}
\usepackage{paralist}
\usepackage{enumitem}
\usepackage[accsupp]{axessibility}

\usepackage[pagebackref=true,breaklinks=true,letterpaper=true,colorlinks,bookmarks=false]{hyperref}

\usepackage[usenames,dvipsnames]{xcolor}

\newcommand{\hj}[1]{\textcolor{black}{#1}}
\newcommand{\hs}[1]{\textcolor{black}{#1}}
\newcommand{\lhs}[1]{\textcolor{black}{#1}}
\newcommand{\dgdg}[1]{\textcolor{black}{#1}}

\newcommand{\ours}[0]{BOSS\xspace}

\usepackage{suffix}
\newcommand\parahead[1]{\vspace{2mm}\noindent\textbf{#1.}\medspace}
\WithSuffix\newcommand\parahead*[1]{\vspace{2mm}\noindent\textbf{#1}\medspace}

\DeclareMathOperator*{\argmax}{arg\,max}
\DeclareMathOperator*{\argmin}{arg\,min}

\iccvfinalcopy 

\def\iccvPaperID{1654} 

\ificcvfinal\fi

\begin{document}

\title{Bayesian Optimization Meets Self-Distillation}


\author{
HyunJae Lee\thanks{Authors contributed equally.} \quad Heon Song$^{*}$ \quad Hyeonsoo Lee$^{*}$ \quad Gi-hyeon Lee \quad Suyeong Park \quad Donggeun Yoo\\
Lunit Inc.\\
{\tt\small \{hjlee, hslee, dgyoo\}@lunit.io,}
{\tt\small \{songheony, lghsigma597, suyeong.park0\}@gmail.com}\\
}

\maketitle
\ificcvfinal\thispagestyle{empty}\fi

\begin{abstract}
Bayesian optimization (BO) has contributed greatly to improving model performance by suggesting promising hyperparameter configurations iteratively based on observations from multiple training trials. However, only partial knowledge (i.e., the measured performances of trained models and their hyperparameter configurations) from previous trials is transferred.
On the other hand, Self-Distillation (SD) only transfers partial knowledge learned by the task model itself.
To fully leverage the various knowledge gained from all training trials, we propose the \ours framework, which combines BO and SD.
\ours suggests promising hyperparameter configurations through BO and carefully selects pre-trained models from previous trials for SD, which are otherwise abandoned in the conventional BO process.
\ours achieves significantly better performance than both BO and SD in a wide range of tasks including general image classification, learning with noisy labels, semi-supervised learning, and medical image analysis tasks.
Our code is available at \href{https://github.com/sooperset/boss}{https://github.com/sooperset/boss}.
\end{abstract}

\section{Introduction}
\label{sec:introduction}

\begin{figure*}
\begin{center}
    \includegraphics[width=0.95\textwidth]{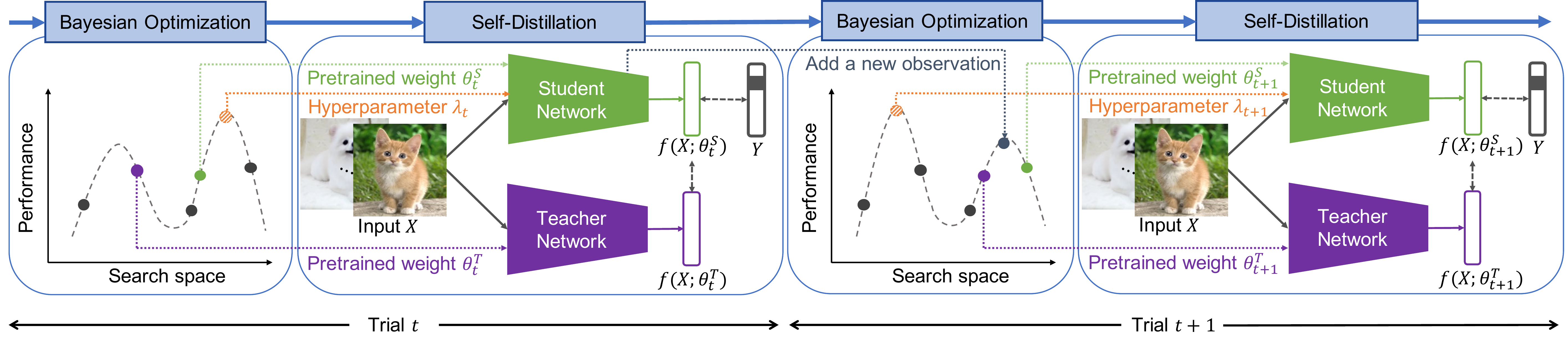}
    \caption{\ours is a novel framework for training models, which fuses the Bayesian Optimization (BO) and Self-Distillation (SD), combining the concept of hyperparameter exploration and knowledge distillation. By performing these steps simply in an alternating manner (left-to-right), we propagate both the conditional probability learned over hyperparameter configurations (depicted with graphs) and the knowledge learned by each task network, resulting in large performance gains in the final model.}
    \label{fig:method}
\end{center}
\vspace{-6mm}
\end{figure*}

Convolutional Neural Networks (CNNs) have achieved remarkable success in a wide range of computer vision applications \cite{chen2017deeplab, lee2019srm, liu2016ssd}. However, their performance is greatly dependent on the choice of hyperparameters \cite{choi2019empirical}. As the optimal hyperparameter configuration is not known a priori, practitioners often explore the hyperparameter space manually to obtain a better configuration. Despite its time-consuming process, it typically results in sub-optimal performance \cite{bergstra2012random}.
Recently, Bayesian optimization (BO) has emerged as a successful approach to hyperparameter optimization, automating the manual tuning effort and pushing the boundaries of performance \cite{bergstra2011algorithms,hutter2011sequential,snoek2012practical}. BO allows for the effective exploration of multi-dimensional search spaces by suggesting promising configurations based on observations. This technique has achieved state-of-the-art performance for training CNNs \cite{bergstra2013making, falkner2018bohb} and has contributed to improving various applications such as AlphaGo \cite{chen2018bayesian}. 

BO is inherently an iterative process in which a probabilistic prior model is fitted using observations of hyperparameter configurations and their corresponding performances \cite{bergstra2011algorithms, hutter2011sequential}. At each iteration, BO suggests the next configuration to evaluate that is most likely to improve performance. After training the network with the suggested configuration, a new observation is retrieved and used to update the probabilistic model. 
However, only partial knowledge (i.e., the measured performances of trained models and their hyperparameter configurations) from previous trials is transferred, and the knowledge learned by the network is discarded.

Self-distillation (SD) can also be viewed as a knowledge transfer method. A recent line of research in SD has demonstrated that transferring knowledge from a previously trained model with the identical capacity can improve the performance of the model \cite{allen2020towards, furlanello2018born, pham2022revisiting}.
If a student network is trained to mimic the feature distribution of a teacher network, then the student could beat the teacher.
Allen-Zhu and Li \cite{allen2020towards} have both theoretically and empirically interpreted this as a similar effect to ensembling various models.
Inspired by this property of SD and the iterative nature of BO, we ended up asking if the knowledge inside the network from the previous trials could be used for the next trials during the BO process.

In this paper, we propose a new framework, \textit{Bayesian Optimization meets Self-diStillation} (\ours), which combines BO and SD to fully leverage the knowledge obtained from previous trials. 
The overall process of \ours is illustrated in \autoref{fig:method}. 
Following the BO process, \ours suggests a hyperparameter configuration based on observations that are most likely to improve the performance. 
After that, it carefully selects pre-trained networks from previous trials for the next round of training with SD, which are otherwise abandoned in the conventional BO process.
This process is performed in an iterative manner, allowing the network to persistently improve upon previous trials.
To the best of our knowledge, this is the first work that leverages the knowledge of the network learned during the BO process. This is not a simple combination of two orthogonal methods but we provide solutions to the problem of how to transfer past knowledge (i.e., model parameters, hyperparameters, and performances) appropriately.
The suggested solution is that (1) not only the teacher but the student should be initialized from the prior knowledge, and (2) they should be initialized from different previous trials to fully exploit the prior knowledge. Our thorough ablation analysis supports this.

We evaluate the effectiveness of \ours with various computer vision tasks, such as object classification~\cite{he2016deep}, learning with noisy labels~\cite{han2018co}, and semi-supervised learning~\cite{arazo2020pseudo}. In addition, we also evaluate it with medical image analysis tasks, including medical image classification and segmentation. Our experimental results demonstrate that \ours significantly improves target model performance, outperforming both BO and SD. Furthermore, we conduct comprehensive analysis and ablation studies to further investigate the behavior of \ours. 

In summary, the main contributions of this paper are:
\begin{enumerate}[nolistsep]
  \item We present \ours framework that fully harnesses the knowledge from various models by leveraging the benefits of both BO and SD. 
  \item Exhaustive evaluation experiments demonstrate the efficacy of our framework, as it results in significant performance improvements across diverse scenarios.
  \item In-depth analysis and ablation studies provide essential insights into how to transfer prior knowledge effectively for CNNs.
\end{enumerate}

\section{Related Work}
\label{sec:related_work}

\parahead{Bayesian Optimization}
Bayesian optimization (BO) is a global optimization method for black-box functions \cite{bergstra2011algorithms}. 
Given a set of hyperparameters $\lambda$ and an objective function $\Phi$ (e.g. accuracy), BO models the conditional probability $p(\Phi(\lambda)|\lambda)$ with a surrogate model. In each iteration, it fits the surrogate model given observations then uses the acquisition function to determine which configuration to evaluate next. 
By suggesting probable candidates based on observations, it has shown remarkable performance on various hyperparameter optimization tasks \cite{bergstra2011algorithms,hutter2011sequential,snoek2012practical}. \hs{While other methods~\cite{rasmussen2006gaussian,bergstra2011algorithms} models $p(\Phi(\lambda)|\lambda)$}, the Tree-structured Parzen Estimator (TPE) \cite{bergstra2011algorithms} models $p(\lambda|\Phi(\lambda))$ by partitioning hyperparameter density function into good and bad groups with respect to their corresponding objective values. It has linear complexity on the number of observations and enables scalable hyperparameter optimization. 
There have been various approaches to improve BO by incorporating prior knowledge. A line of works aims to transfer knowledge learned from other datasets through meta-learning. Multi-objective TPE \cite{watanabe2022multi} extends TPE to transfer knowledge from other tasks considering the similarity between tasks. Surrogate-Based Collaborative Tuning \cite{bardenet2013collaborative} leverages the knowledge from various datasets by integrating the surrogate ranking algorithm into BO.
On the other hand, Prior-guided Bayesian Optimization \cite{souza2020prior} allows domain experts to transfer their knowledge into BO in the form of priors.
Our \hs{framework} is different from these methods in that these utilize \hs{prior knowledge from external sources} to better estimate the surrogate model, while we utilize the prior knowledge from \hs{the given task} to enhance the training of the \dgdg{target} model. It means that these methods can be applied orthogonally to our framework. \par

\parahead{Self-Distillation}
Knowledge distillation (KD) is a model compression method that involves transferring the knowledge of a large teacher model to a small student model while maintaining performance.
The original work by Hinton et al.~\cite{hinton2015distilling} proposed distilling knowledge by matching the softmax distribution of the teacher and student models.
Since then, various methods have been introduced to improve the knowledge transfer process. 
Self-Distillation (SD) is a special form of KD where the teacher and student networks have identical architecture. 
Born-Again Networks (BAN) \cite{furlanello2018born} demonstrated that when training the student to match the output distribution of the teacher with the identical architecture, it could outperform the teacher. Furthermore, they showed that performing multiple rounds of BAN could further improve the performance where the trained student is set to be a new teacher in the following round.
The effectiveness of SD has been theoretically explained by the ``multi-view'' hypothesis introduced by Allen-Zhu and Li, who showed that self-distillation performs an implicit ensemble of various models \cite{allen2020towards}. Empirical evidence from Pham et al. \cite{pham2022revisiting} suggests that SD encourages the student to find flatter minima, leading to better generalization.
In this work, we identify that SD can be an effective method for propagating the task knowledge learned in early stages of BO, to late stages of BO. This combination of the SD and BO processes is key to yielding a high-performing model, which we validate experimentally.


\section{Method}
\label{sec:method}

In this section, we introduce \ours which \hs{combines BO and SD} to leverage the knowledge learned in the training trials.
We first provide a brief introduction to BO and SD then introduce \ours in detail.

\subsection{Bayesian Optimization}
BO is an iterative process that aims to optimize an objective function $\Phi$ with respect to a set of hyperparameters $\lambda$ \cite{hutter2011sequential}.
At each iteration $t$, BO builds a surrogate model to approximate $\Phi$ based on the observations from the previous iterations, denoted by $\Lambda_{t-1} = \{(\lambda_0, \Phi(\lambda_0)), \ldots, (\lambda_{t-1}, \Phi(\lambda_{t-1}))\}$.
 BO then uses an acquisition function $\mu(\lambda|\Lambda_{t-1})$ to select the next hyperparameter configuration $\lambda_t$ to evaluate. 
 The acquisition function balances exploration and exploitation, with the most common choice being the Expected Improvement (EI) \cite{jones1998efficient}:
\begin{align}
\mu(\lambda | \Lambda) = \mathop{\mathbb{E}}(\max\{0, \Phi(\lambda) - \Phi_{\max}\} | \Lambda),
\label{equation:ei}
\end{align}
where $\Phi_{\max}$ is the best observed function value. The hyperparameter configuration at iteration $t$ is selected as $\lambda_{t} = \argmax_\lambda \mu(\lambda|\Lambda_{t-1})$.
After evaluating $\Phi(\lambda_t)$, the new observation is added to the previous ones, $\Lambda_t=\Lambda_{t-1}\cup\{(\lambda_t,\Phi(\lambda_t))\}$, and the surrogate model is updated. 

While many surrogate models including the Gaussian process (GP) directly models $p(\Phi(\lambda)|\lambda)$, 
the Tree-structured Parzen Estimator (TPE) \cite{bergstra2011algorithms} models $p(\lambda|\Phi(\lambda))$ which is defined by two functions:
\begin{align}
p(\lambda|\Phi(\lambda))\;=\; \left\{
        \begin{array}{rcl}
        l(\lambda), & \hj{\Phi}(\lambda)\;>\;\Phi^*\\
        g(\lambda), & \hj{\Phi}(\lambda)\;\leq\;\Phi^*\\
        \end{array} \right. .
\end{align}
Here, $l(\lambda)$ is the ``good'' density formed by observations that the performance was higher than a threshold $\Phi^*$\footnote{In general, $\Phi^*$ is set to be some quantile of objective values $p(\Phi(\lambda)>\Phi^*)=\gamma$ \cite{bergstra2011algorithms}.}, $g(\lambda)$ is the ``bad'' density formed by the remaining observations.
Bergstra et al. \cite{bergstra2011algorithms} claims that maximizing EI is proportional to maximizing the ratio of $l(\lambda) / g(\lambda)$. 
Hence, on iteration $t$, TPE suggests a configuration $\lambda_t$ that maximizes this ratio. Despite its simplicity, TPE has outperformed traditional surrogate models such as GP \cite{rasmussen2006gaussian}.

\subsection{Self-Distillation}
Furlanello et al.~\cite{furlanello2018born} proposed the Self-Distillation (SD) technique for transferring knowledge from a teacher network to a student network with the same architecture. Given a neural network $f$ initialized with random parameters ${\theta}$, SD first obtains the parameters of the teacher model $\theta^T$ 
by minimizing the task loss $\mathcal{L}_{gt}(f(X;\theta),Y)$ with respect to $\theta$. SD then obtains the parameters of the student model $\theta^S$ by minimizing a loss with respect to $\theta$ that balances the task loss $\mathcal{L}_{gt}$ and the distillation loss $\mathcal{L}_{dt}$ as follows:
\begin{equation}
\alpha\mathcal{L}_{gt}(f(X;\theta),Y) + (1 - \alpha)\mathcal{L}_{dt}(f(X;\theta), f(X;\theta^T)),
\label{equation:sd}
\end{equation}
where $\alpha$ is a hyperparameter for balancing the two losses. \par
In various image classification tasks, the cross-entropy loss is used to define the task loss $\mathcal{L}_{gt}$ as $\mathcal{L}_{gt}(f(X;\theta), Y) = -\sum Y \log \sigma (f(X;\theta))$, where $\sigma$ is the softmax function. For the distillation loss, in conventional self-distillation, $\mathcal{L}_{dt}$ is defined using the Kullback-Leibler (KL) divergence loss. However, recent work~\cite{kim2021comparing} has shown that the mean squared error (MSE) between logits from the student and teacher outperforms the KL divergence loss. The MSE distillation loss is defined as $ \mathcal{L}_{dt}(f(X;\theta), f(X;\theta^T)) = ||f(X;\theta^T) - f(X;\theta) ||_2^2$.\par

\setlength{\textfloatsep}{13pt}
\begin{algorithm}[t]
\SetAlgoLined
\textbf{Input:} Training data $(X, Y)$; number of total trials~$N$; number of warm-up trials $W$; number of candidates $K$; neural network $f$; balance parameter $\alpha$. \\
\textbf{Initialize:} Sets of observations $\Lambda_0,\Lambda'_W = \emptyset$; set of parameters $\Theta_0 = \emptyset$. \\
\textbf{Phase 1: Warm-up}\\
\For{$t = 1, 2, \ldots, W$}{
Draw $\theta_t$ from the standard normal distribution\\
Find $\lambda_t$ that maximizes $\mu(\lambda | \Lambda_{t-1})$\\
Obtain $\theta_t^*$ for $\lambda_t$ that minimizes $\mathcal{L}_{gt}$ w.r.t $\theta_t$:
\begin{equation*}
\theta^*_t=\argmin_{\theta_t} \mathcal{L}_{gt}(f(X;\theta_t),Y)
\end{equation*}\\
Update $\Lambda$: $\Lambda_t = \Lambda_{t-1} \cup \{(\lambda_t,\Phi(\lambda_t))\}$\\
Update $\Theta$: $\Theta_t = \Theta_{t-1} \cup \{\theta_t^*\}$.
}
\textbf{Phase 2: \ours Training}
\\
\For{$t = W+1, W+2, \ldots, N$}{
Randomly select $\theta_t^{S}, \theta_t^{T} \in \textsc{top-}K(\Theta_{t-1})$ \\
Find $\lambda_t$ that maximizes $\mu(\lambda | \Lambda'_{t-1})$\\
Obtain $\theta_t^*$ for $\lambda_t$ that minimizes Eq.~\ref{equation:sd} w.r.t $\theta_t^S$:
\begin{align*}
\theta^*_t=\argmin_{\theta_t^S} &~\alpha\mathcal{L}_{gt}(f(X;\theta_t^{S}),Y) \\
& + (1-\alpha)\mathcal{L}_{dt}(f(X;\theta_t^{S}),f(X;\theta_t^{T})) 
\end{align*}\\
Update $\Lambda'$: $\Lambda'_t = \Lambda'_{t-1} \cup \{(\lambda_t,\Phi(\lambda_t))\}$\\
Update $\Theta$: $\Theta_t = \Theta_{t-1} \cup \{\theta_t^*\}$.
}
\textbf{Output:} $\theta_\text{best} \in \Theta_N$ showing the best performance.
\caption{Optimization Process for \ours}
\label{algo:lpk}
\end{algorithm}
\subsection{\ours Framework}

\ours integrates SD into the BO process in order to incorporate the knowledge learned in former trials. While the proposed framework is applicable to various BO and SD methods (see \autoref{sec:analysis}), we opt for TPE \cite{bergstra2011algorithms} due to its low computational complexity and scalability to diverse search spaces. In addition, we utilize MSE for our distillation loss as it has been shown to yield better performance than KL divergence while requiring fewer hyperparameters \cite{kim2021comparing}. The overall procedure of \ours is summarized in \autoref{algo:lpk}. 

In order to perform SD, a teacher network is required to train a student network. However, the absence of any network, in the beginning, poses a cold start problem. 
To address this issue, a warm-up phase is introduced, which is similar to the regular BO process but with the difference that the trained CNNs are recorded for use in subsequent phases.
In the warm-up phase, the neural network is initialized with random parameters $\theta$ and a hyperparameter $\lambda$ is suggested using the acquisition function $\lambda = {\argmax}_\lambda \mu(\lambda | \Lambda)$ in \autoref{equation:ei}.
Training images $X$ and their corresponding labels $Y$ are then used to obtain new parameters $\theta^*$ for $\lambda$ by minimizing the task loss $\mathcal{L}_{gt}$. Any task-specific loss can be used such as the cross-entropy loss used in image classification tasks. After training, the performance of the neural network $\Phi(\lambda)$ is computed and the observation $(\lambda, \Phi(\lambda))$ is added to a set of observations $\Lambda$. Also, the obtained parameters $\theta^*$ are added to a set of parameters $\Theta$.

In the second phase where \ours training is performed, an empty set is initialized as a new observation set $\Lambda'$, because the training scheme differs from the warm-up phase and the optimal configuration is likely to be different.
Given the set of parameters $\Theta$ and \dgdg{the pre-defined} number of candidates $K$, \dgdg{the top-$K$ candidates are selected according to the performances in $\Theta$. Among them, two are randomly chosen again to define a student and a teacher.}
\dgdg{We initialize the teacher $\theta^T$ and student $\theta^S$ with different parameters as we expect some benefit from aggregating different knowledge from different networks.}
\dgdg{The effect of this scheme} is elaborated further in \autoref{ssec:pretrained}.
After \dgdg{the initialization, the} hyperparameter $\lambda$ is suggested \dgdg{in the same procedure} as in the warm-up phase. 
Unlike the warm-up phase, $\theta$ is trained to minimize the loss in \autoref{equation:sd}. \dgdg{Once} training is complete, the performance and trained parameters are added to $\Lambda'$ and $\Theta$, respectively. \dgdg{This procedure is repeated until the budget $N$ is exhausted.}
\hs{Despite potential concerns that the utilization of pre-trained weights could interfere with BO's modeling, the previous study~\cite{kim2017learning} has demonstrated that BO is sufficiently robust to the variability of initial weights by randomly initializing them at each trial.}
\par

As \ours training proceeds, the top-$K$ models are updated with the models that incorporate the knowledge of previous trials. These models are then utilized to transfer the knowledge in the follow-up iterations. By iteratively updating the candidates with enhanced networks, \ours consistently improves performance upon previous trials.

\section{Experiments}
\label{sec:experiments}

In this section, we present a comprehensive evaluation of \ours on a wide range of problems and datasets.
We compare it with several other methods in our comprehensive evaluation, including 
Baseline (conventional training with hyperparameters tuned by human experts), SD (Self-Distillation with MSE distillation loss~\cite{kim2021comparing}), Random (Random search for hyperparameter optimization~\cite{bergstra2012random}), and BO (Bayesian Optimization with TPE~\cite{bergstra2011algorithms}).
Motivated by the setting in \cite{lee2023improving}, we define our search space to include learning rate $l$, momentum $1 - m$, weight decay $w$ and batch size $b$ where $l, m$ are sampled from $\log[10^{-3}, 1]$, $w$ from $\log[10^{-5}, 10^{-2}]$, and $b$ from $[64, 256]$.
We use an implementation of TPE from the Optuna framework~\cite{optuna_2019} and conduct 128 trials with parallel execution of 8 trials. 
$K$ and $W$ are set to 8 and 32 respectively which will be further explored in \autoref{sec:ablation}. 

In order to ensure a fair comparison with \ours, we conduct multiple rounds of SD until it reaches saturation and report the highest accuracy obtained among all the rounds.
Specifically, we continue the SD process until there is no further improvement in performance for three consecutive rounds. 
We set the hyperparameter $\alpha=0.5$ for SD, which we find to yield the best performance among $\alpha \in \{0.25, 0.5, 0.75\}$ on CIFAR-100 \cite{krizhevsky2009learning} with the VGG-16 \cite{simonyan2014very} architecture. However, we empirically observe that the final performance does not vary significantly when the SD process is continued until saturation.

\subsection{Object Classification}
\label{sec:classification}
We first evaluate the effectiveness of \ours in object classification tasks. Publicly available implementations\footnote{\url{https://github.com/bearpaw/pytorch-classification}} of classification networks and training procedures are utilized to fairly compare it with the human baseline.
We use three standard datasets: CIFAR-10 and CIFAR-100 \cite{krizhevsky2009learning}, which are comprised of 50,000 training and 10,000 test images of 10 and 100 object classes, respectively, and Tiny ImageNet \cite{russakovsky2015imagenet}, which includes 100,000 training and 10,000 validation images across 200 object classes.
Consistent with previous works \cite{lee2023improving, yun2020regularizing}, we resize the images in the Tiny ImageNet dataset to 32x32 pixels to align with the training procedure of the CIFAR datasets.

In our experiments, the networks are trained using the stochastic gradient descent (SGD) optimizer for 164 epochs on a single GPU. The baseline hyperparameters include a learning rate of 0.1, a weight decay of 5e-4, a momentum of 0.9, and a batch size of 128. We follow the standard practice \cite{he2016deep} for training-time data augmentation by zero-padding each image with 4 pixels, randomly cropping it to the original size, and performing evaluations on the original images.

\begin{table}[t]
\centering
\addtolength{\tabcolsep}{-1pt}
\begin{tabular}{lccc}
\toprule
Method & CIFAR-10 & CIFAR-100 & Tiny-ImgNet \\
\midrule
Baseline & 93.75$_{\pm0.18}$ & 74.43$_{\pm0.49}$ & 53.33$_{\pm0.49}$ \\
SD & 94.21$_{\pm0.24}$ & 76.08$_{\pm0.28}$ & 56.46$_{\pm0.60}$  \\ 
\midrule
Random & 94.06$_{\pm0.39}$ & 75.66$_{\pm1.30}$ & 54.72$_{\pm0.66}$ \\ 
BO & 94.52$_{\pm0.20}$ & 76.48$_{\pm0.17}$ & 55.39$_{\pm0.31}$  \\ 
\ours (Ours) & \textbf{94.98$_{\pm0.19}$} & \textbf{77.69$_{\pm0.15}$} & \textbf{58.55$_{\pm0.35}$}  \\ 
\bottomrule
\end{tabular}
\vspace{2pt}
\caption{Top-1 accuracy (\%) on CIFAR10/100 and Tiny-ImageNet with VGG-16. The reported results are the average and the 95\% confidence interval over 5 repetitions.}
\vspace{-1mm}
\label{table:various_datasets}
\end{table}

\parahead{Performance on different datasets}
We train the VGG-16 architecture \cite{simonyan2014very} on the aforementioned three datasets. \autoref{table:various_datasets} compares the top-1 accuracy with the average and the 95\% confidence interval over 5 repetitions. It demonstrates that \ours exhibits considerably higher accuracy than other methods across all tested datasets.
While random search succeeds to improve the performance of the baseline, BO further boosts the performance by adaptively suggesting probable configurations. SD also achieves enhanced performance compared to the baseline as expected. 
However, the effectiveness of SD and BO varies across datasets. On the other hand, \ours consistently improves the performance by a large margin, leveraging the advantages of both methods.

\begin{figure}[t]
\begin{center}
\label{fig:cifar100}
\includegraphics[width=0.91\linewidth]{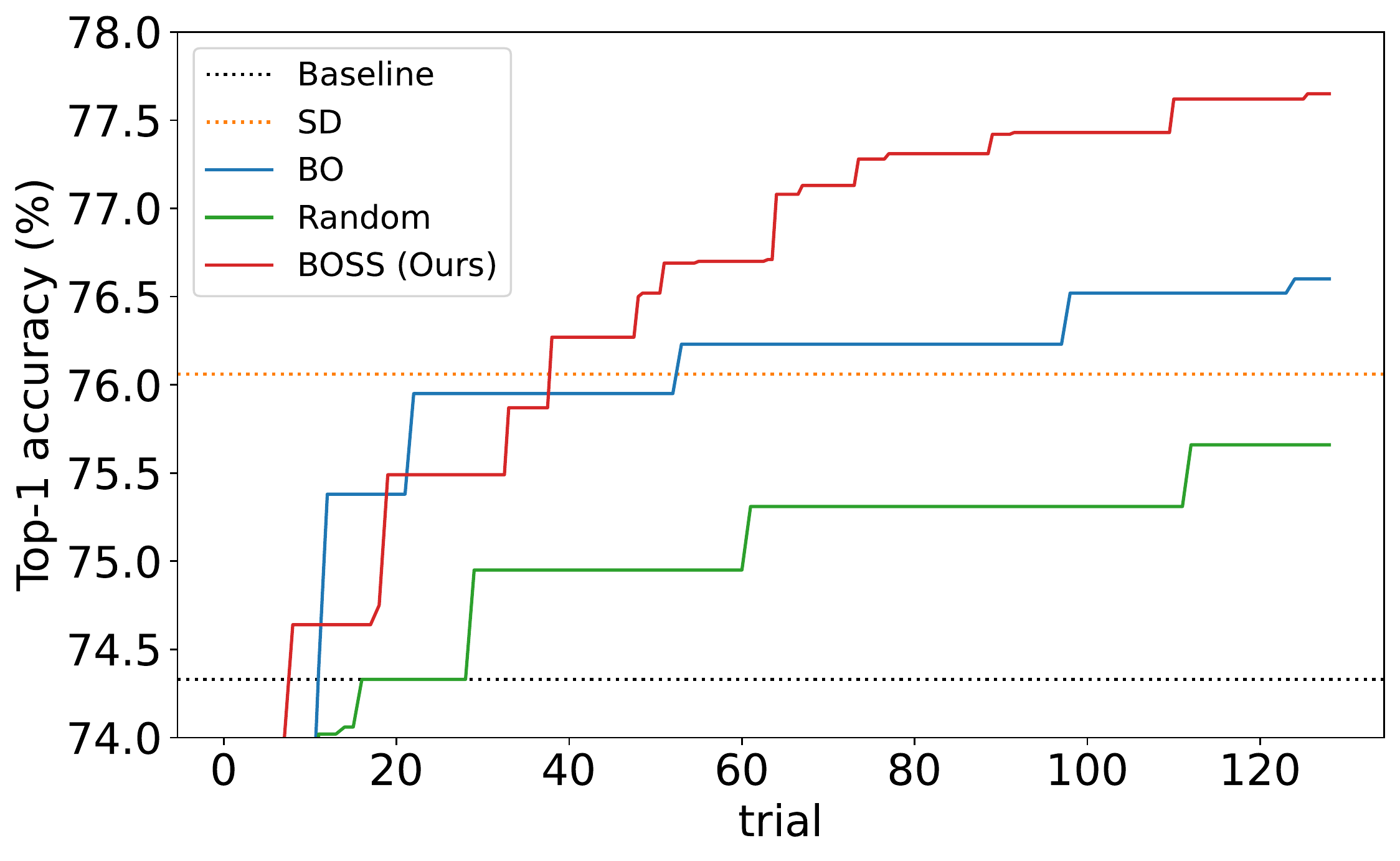}
\caption{Performance curves of various methods on CIFAR-100 with VGG-16 architecture. 
}
\label{fig:comparison}
\end{center}
\vspace{-4mm}
\end{figure}
\parahead{Comparison of diverse methods}
We compare the performance of \ours with other methods in terms of the number of trials.
\autoref{fig:comparison} illustrates the performance curves of Random, BO and \ours along with the Baseline and SD on CIFAR-100 with VGG-16. 
Throughout the optimization process -- except for the initial warm-up phase -- \ours consistently exhibits superior performance over other methods. 
Initially, all methods display similar performance due to the lack of adequate observations, and BO outperforms random search by exploring promising configurations using previous observations with an increasing number of trials.
However, after a certain point, the performance improves at a slow pace. 
In contrast, \ours persistently improves upon previous trials by leveraging prior knowledge as a stepping stone to push the performance boundary.
This suggests that \ours's efficacy is not easily saturated, making it worthwhile for future investigation.

\begin{table}[t]
\centering
\addtolength{\tabcolsep}{-2.3pt}
\begin{tabular}{lcccc}
\toprule
Method & AlexNet & VGG-11  & ResNet-20 & ResNet-56 \\
\midrule
Baseline & 45.73 & 71.21 & 68.92 & 72.04 \\ 
SD & 46.43 & 73.27 & 69.89  & 75.23 \\ 
\midrule
Random & 46.07 & 72.18 & 68.58 & 72.19 \\ 
BO & 46.86 & 73.70 & 69.84 & 74.82 \\ 
\ours (Ours) & \textbf{50.61} & \textbf{75.61} & \textbf{72.22} & \textbf{76.26}  \\ 
\bottomrule
\end{tabular}
\vspace{2pt}
\caption{Top-1 accuracy (\%) on CIFAR-100 with different network architectures.}
\label{table:various_archs}
\vspace{-2mm}
\end{table}

\parahead{Scalability to various architectures}
We demonstrate the scalability of \ours with respect to a variety of CNN architectures including AlexNet \cite{krizhevsky2012imagenet}, VGG-11 \cite{simonyan2014very} and ResNet-20/56 \cite{he2016deep} on the CIFAR-100 dataset. 
As shown in \autoref{table:various_archs}, \ours consistently outperforms all other methods across all tested architectures. 
Notably, even for small networks such as ResNet-20, \ours achieves superior performance compared to the baseline of larger ResNet-56 with more than three times as many parameters. This finding suggests that \ours enables efficient network design by achieving better performance with fewer parameters. 
Moreover, it further brings additional performance improvement to large networks, implying that \ours is not simply substituting the performance gain by increasing the capacity of the CNN architecture.

\parahead{Performance with enhanced baseline} 
\begin{table}[t]
\centering
\begin{tabular}{lcccc}
\toprule
 & Baseline & SD & BO  & \ours \\
\midrule
CIFAR-10 & 95.01 & 95.65 & 95.79 & \textbf{96.77} \\ 
CIFAR-100 & 77.09 & 78.44 & 78.07 & \textbf{80.81} \\ 
\bottomrule
\end{tabular}
\vspace{6pt}
\caption{Top-1 accuracy (\%) on CIFAR-10/100 with advanced augmentation methods using VGG-16 architecture.}
\label{table:more_augs}
\vspace{-1mm}
\end{table}

Pham et al.~\cite{pham2022revisiting} criticize the existing literature on SD, saying that the performance of reported baselines is not fully saturated. In such cases, the gains achieved through distillation may be invalidated if the baseline is better optimized. 
Following their suggestion, we incorporate the advanced data augmentations of AutoAugment \cite{cubuk2019autoaugment} and Cutout \cite{devries2017improved}. 
We train VGG-16 on CIFAR-10/100 using the aforementioned regularizations and report the top-1 validation accuracy in \autoref{table:more_augs}. As anticipated, the updated data augmentations yield a substantial performance improvement over those in \autoref{table:various_datasets}. 
Nonetheless, \ours consistently outperforms other methods by a considerable margin, indicating that the improvements achieved by \ours are not hindered by better training schemes such as advanced regularization methods.

\subsection{Learning with Noisy Labels}
\label{sec:noise}

\begin{table}[t]
\centering
\begin{tabular}{lcccc}
\toprule
\multirow{2}{*}{Method} & \multicolumn{2}{c}{Symmetric} & \multicolumn{2}{c}{Asymmetric} \\
\cmidrule(lr){2-3} \cmidrule(lr){4-5}
& 20\% & 40\% & 20\% & 40\% \\
\midrule
Baseline & 64.79 & 53.34 & 63.09 & 47.44 \\
SD & 68.93 & 59.58 & 65.73 & 49.12 \\
\midrule
BO & 65.47 & 57.52 & 67.52 & 56.48 \\
\ours (Ours) & \textbf{70.34} & \textbf{65.50} & \textbf{72.82} & \textbf{67.38} \\ 
\bottomrule
\end{tabular}
\vspace{6pt}
\caption{Comparison with Different Noise Rates on CIFAR-100}
\label{table:label_noise}
\vspace{-2mm}
\end{table}

We further demonstrate the benefit of \ours in noisy label settings where incorrect labels exist in training data.
In real-world scenarios, the presence of noisy labels
often hinder the performance of CNN models \cite{song2022learning}. To address this challenge, previous research~\cite{allen2020towards, kaplun2022knowledge} has demonstrated that employing distillation methods could mitigate the impact of noisy labels by incorporating distinct aspects from multiple teachers to the student.
\ours builds upon this approach by iteratively transferring different knowledge obtained from previous trials. 

We conduct experiments to show the robustness of \ours by training VGG-16 on CIFAR-100 following the training procedure in \autoref{sec:classification}. \autoref{table:label_noise} shows the top-1 accuracy with respect to varying noise ratios and types where symmetric noise refers to the situation where all labels are mixed with the same probability of noise, while asymmetric noise flips labels based on similar class pairs.
As expected, the performance of the baseline drops significantly compared to training with the clean dataset in \autoref{table:various_datasets}. While both SD and BO improve the performance of the baseline to some extent, they \lhs{are sensitive to noise and exhibit performance decreases under large noise conditions.} On the other hand, \ours consistently achieves significant performance improvement regardless of noise levels and types which suggests that the negative influence of label noise can be greatly mitigated by the proposed framework.

\subsection{Semi-Supervised Learning}
\label{sec:semi}


\lhs{We evaluate \ours in the context of semi-supervised learning (SSL), where a large amount of unlabeled data is available.}
\hj{Previous works~\cite{arazo2020pseudo, radosavovic2018data, tarvainen2017mean, xie2020self} have demonstrated that transferring past knowledge via self-training is an effective approach for utilizing unlabeled data. However, it generally does not exploit the knowledge from various models. We investigate whether \ours could further enhance the performance by incorporating the knowledge of multiple high-performing models yielded during BO process.}

\hj{We leverage one of the state-of-the-art SSL method \cite{arazo2020pseudo} as a baseline} and adopt the official implementation\footnote{\url{https://github.com/EricArazo/PseudoLabeling}}. We opt for 13-CNN architecture \cite{athiwaratkun2018there} which is mainly explored in the paper. \autoref{table:semi_supervised} presents the top-1 accuracy of CIFAR-100 dataset with 4,000 and 10,000 labeled data. It is worth noting that the baseline is trained with carefully tuned hyperparameters, advanced regularization techniques, and long training iterations. Nevertheless, \ours consistently improves the accuracy with a meaningful margin.
\hj{This result indicates that \ours could generate a positive synergy with existing SSL algorithm, complementing it by leveraging the knowledge from various models.}

\begin{table}[t]
\centering
\addtolength{\tabcolsep}{-2.3pt}
\begin{tabular}{lcccc}
\toprule
 & Baseline & SD & BO  & \ours \\
\midrule
4,000 labels & 63.11 & 63.80 & 64.37 & \textbf{65.93} \\ 
10,000 labels & 67.78 & 69.18 & 69.31 & \textbf{70.93} \\ 
\bottomrule
\end{tabular}
\vspace{6pt}
\caption{Top-1 accuracy (\%) of semi-supervised learning on CIFAR-100 with varying numbers of labeled data.}
\label{table:semi_supervised}
\vspace{-2mm}
\end{table}

\subsection{Medical Image Analysis Tasks}
\label{sec:medical}


The significance of medical image analysis tasks lies in their direct impact on patient outcomes, making their performance a critical concern. One approach to enhance performance in these tasks is to employ multiple models, like model ensembles~\cite{hosni2019reviewing,hameed2020breast}. However, this can be limited by resource constraints in real-world environments such as hospitals. To tackle this challenge, researchers have turned to HPO techniques such as BO, despite their substantial costs, to improve model performance~\cite{ritter2019hyperparameter,gao2020disease}. Therefore, in this section, we show the efficacy of \ours on two critical medical image analysis tasks.
\par
The first task we evaluate is breast cancer classification using mammograms, which are x-ray images of the breast used for early cancer detection. Accurate mammographic breast cancer classification is essential, being the primary detection method and the second most common cancer in women.
We use two datasets for breast cancer classification with mammograms: the publicly available Chinese Mammography Database~\cite{cui2021chinese} (CMMD, 826 exams), and the in-house dataset sourced from the European Union (INH, 6,994 exams). We split each dataset into training sets and validation sets in a 7:3 ratio.
We use ResNet-34 to classify mammograms into two categories: cancer or non-cancer. The mammograms are resized to 1920$\times$1280 and applied with various geometric and photometric augmentations, such as translate, rotate, shear, flip, and brightness/contrast adjustment.
The model's performance is evaluated using the Area under the ROC Curve (AUC)\footnote{The exam-level AUC~\cite{salim2020external} is calculated, and only exams containing a complete set of 4-view images are considered.}.
\par



\begin{table}[t]
\centering
\setlength{\tabcolsep}{4pt}
\begin{tabular}{l@{\hspace{5mm}}cc|cc}
\toprule
\multirow{2}{*}{Method} & \multicolumn{2}{c|}{Breast (AUC)} & \multicolumn{2}{c}{Nuclei (mPQ)} \\
\cmidrule(lr){2-3} \cmidrule(lr){4-5}
& CMMD & INH & CoNSeP & CPM-17 \\
\midrule
Baseline & 79.22 & 87.55 & 51.67 & 68.03 \\
BO & 81.61 & 87.95 & 52.88 & 69.83 \\
\ours (Ours) & \textbf{85.81} & \textbf{89.57} & \textbf{53.41} & \textbf{71.66} \\ 
\bottomrule
\end{tabular}
\vspace{3pt}
\caption{Performance on medical image analysis tasks}
\label{table:medical_imaging}
\vspace{-2mm}
\end{table}

Next, we address the task of nuclei instance segmentation, which involves identifying and segmenting individual nuclei in histopathology images. This task is particularly crucial in medical research, as it can assist in the diagnosis and treatment of diseases like cancer. To evaluate the proposed method, we conduct experiments on two datasets: 
CoNSeP~\cite{graham2019hover} (24,319 instances) and CPM-17~\cite{vu2019methods} (7,570 instances). To ensure a fair comparison, we train the HoVer-Net~\cite{graham2019hover} architecture, which is a standard architecture in the task, using the official repository\footnote{\url{https://github.com/vqdang/hover_net}} and training schemes. The hyperparameter configuration recommended in the original implementation serves as the baseline. We use the multi-class Panoptic Quality (mPQ)~\cite{graham2019hover} metric 
to evaluate the performance of the model, with higher scores indicating better performance.

\autoref{table:medical_imaging} shows that \ours outperforms both the baseline and BO methods, achieving enhanced performance on both tasks.
This demonstrates that \ours can efficiently 
utilize the limited medical image data without any additional overhead at test time via an implicit ensemble of multiple models and superior hyperparameter configuration. It is consistent with the previous findings that have shown model ensembles~\cite{yang2010review} and hyperparameter optimization~\cite{lee2023improving} to be effective when data is limited. Moreover, the result further shows that the advantage of \ours is not restricted to classification tasks but could be readily extended to instance segmentation tasks as well. Further extension of \ours to other tasks such as object detection \cite{liu2016ssd} or generative modeling \cite{rombach2022high} remains as future work.

\section{Ablation Study and Analysis}
\label{sec:analysis}

This section presents ablation studies and analytical experiments to investigate the design choices of \ours algorithm. 
We conduct these studies using the same experimental setup as the object classification experiment on VGG-16 with CIFAR-100 presented in \autoref{sec:classification}.


\begin{figure}[t]
\begin{center}
\includegraphics[width=\linewidth]{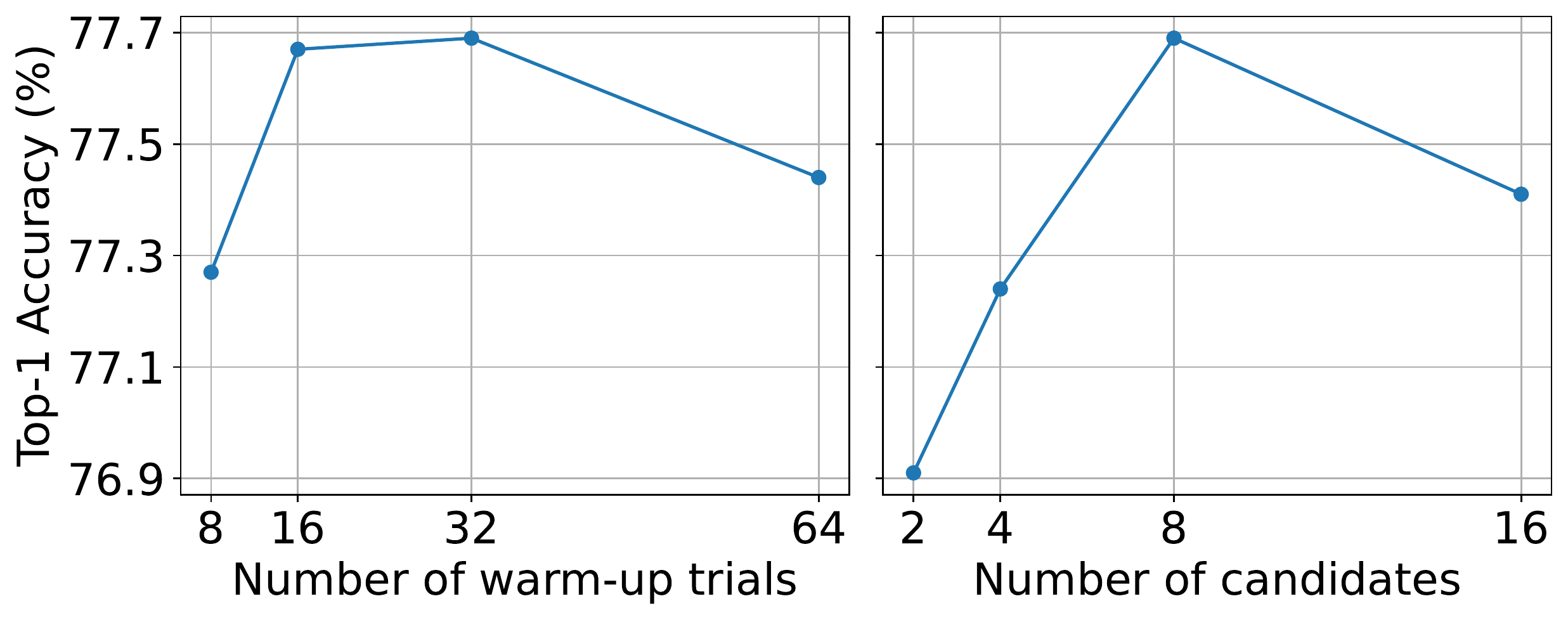}
\caption{Top-1 accuracy of \ours on CIFAR-100 with varying the number of (left) warm-up trials and (right) candidates.
\label{fig:fablation}}
\end{center}
\vspace{-4mm}
\end{figure}

\begin{table}[t]
\centering
\begin{tabular}{cccc}
\toprule
BO & PT Student & PT Teacher & Top-1 Accuracy \\
\midrule
\ding{51} & \ding{55} & \ding{55} & 76.48 \\ 
\ding{51} & \ding{51} & \ding{55} & 76.93 \\
\ding{51} & \ding{55} & \ding{51} & 77.18 \\ 
\ding{51} & \ding{51} & \ding{51} & \textbf{77.69}\\ 
\bottomrule
\end{tabular}
\vspace{6pt}
\caption{Comparison of different choices for utilizing pretrained weights on CIFAR-100 with VGG-16. "PT" stands for "Pre-trained".}  
\label{table:various-pretrained}
\vspace{-2mm}
\end{table}
\subsection{Ablation Study}
\label{sec:ablation}
We first examine the impact of the number of warm-up trials $W$ on the performance of \ours.
As BO typically requires a sufficient number of trials to achieve good performance, pretrained models might exhibit low performance when warm up trials are not enough. On the other hand, too many warm-up trials may leave insufficient budget for \ours training.
\autoref{fig:fablation} shows that \ours performs best with $W=32$, but the performance is not sensitive to $W$. This suggests that the warm-up phase and \ours phase can complement each other, even with small or large $W$.
We further conduct an ablation study on the number of candidates $K$ from which the teacher and student networks are randomly selected. 
\autoref{fig:fablation} also shows that \ours achieves the best performance when $K=8$. Choosing a smaller $K$ might limit the performance by not leveraging sufficient knowledge, while selecting a larger $K$ might reduce performance by including weak knowledge from the teacher network.

\ours leverages pretrained weights for both teacher and student networks to capitalize on past knowledge. 
In~\autoref{table:various-pretrained}, we explore different approaches to utilizing pretrained weights during the BO process.
When the pretrained weights are loaded only on the student, distillation is not applied. It can be considered as a form of warm restart which has been demonstrated to enhance the performance of DNNs~\cite{loshchilov2017sgdr}. 
Employing pretrained weights for either the student or teacher network results in improved performance compared to standard BO.
Furthermore, utilizing pretrained weights for both the teacher and student networks leads to even greater performance gains. This indicates that the knowledge of teacher and student could generate a positive synergy.


\begin{figure}
\begin{center}
    \includegraphics[width=0.32\textwidth]{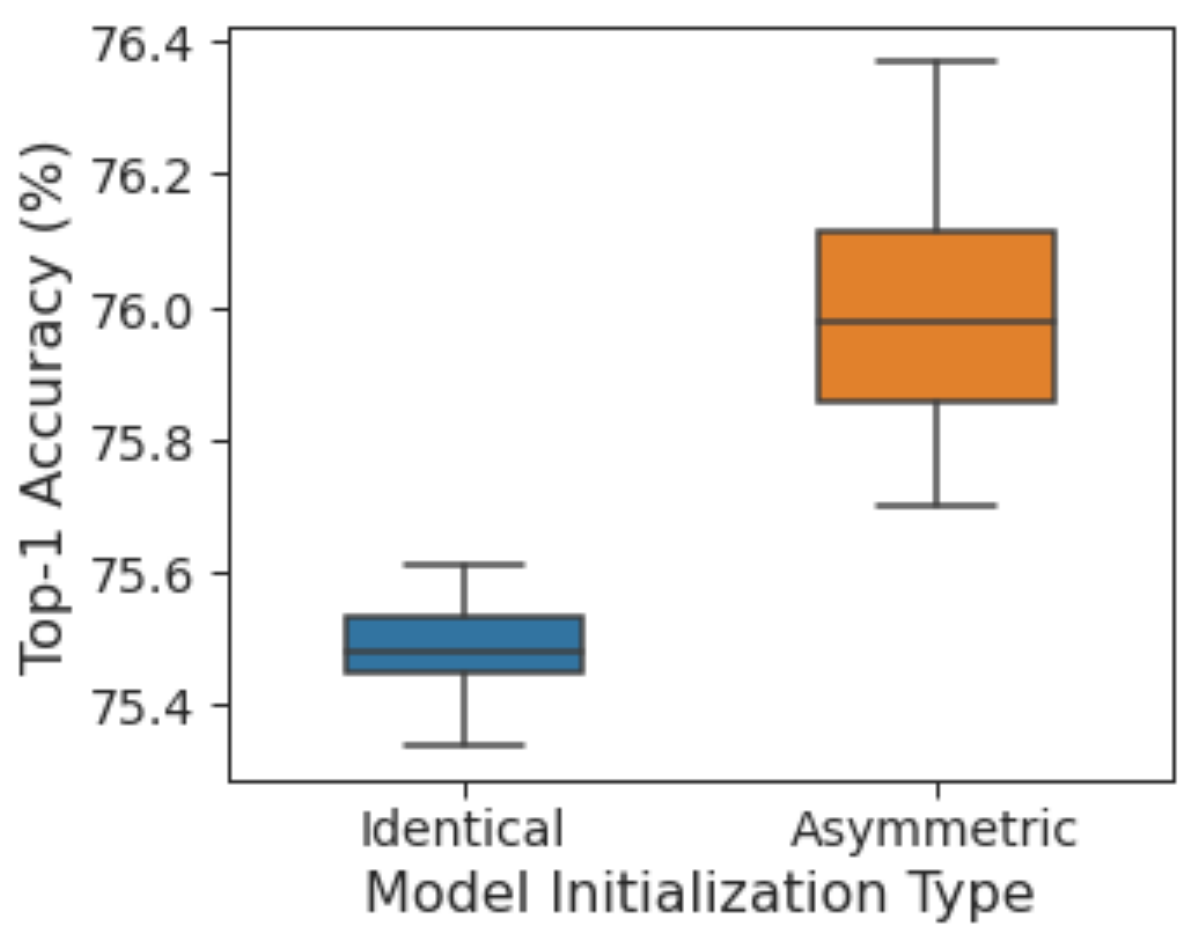}
    \caption{Boxplots of Top-1 accuracy on CIFAR-100 with VGG-16 between identical and
asymmetric models. Identical models have the same initialization weights for both teacher and student, while asymmetric models have different initialization weights.}
    \label{fig:initialization}
\end{center}
\vspace{-2mm}
\end{figure}
\subsection{Effect of Pretrained Weight}
\label{ssec:pretrained}
In light of our observations that utilizing pretrained weights for both student and teacher networks leads to a improved performance, we note that the Noisy Student \cite{xie2020self} reported no improvement in performance when initializing both models simultaneously. They claim that this approach may sometimes lead to getting stuck in a local optima, resulting in inferior performance compared to training the student model from scratch.
We hypothesize that this contrasting result could be due to the fact that they used the same pretrained weight for both teacher and student. Since the student model has already inherited all the knowledge that could be learned from the teacher, distilling knowledge with the teacher may not bring additional performance gains.

A recent study by Allen-Zhu and Li \cite{allen2020towards} revealed that different networks trained with distinct random seeds learn different knowledge. Motivated by this observation, we train VGG-16 on CIFAR-100 with 8 different seeds following the same procedure as the baseline in \autoref{sec:classification} and utilize them to initialize teacher and student networks. For all possible combinations of teacher and student networks, we conducted a single-round self-distillation, comparing the top-1 accuracy of the trained models between identical and asymmetric initialization.
The identical models refer to initializing both teacher and student with the same pretrained model, while asymmetric models refer to initializing them with different models.
The results are presented in \autoref{fig:initialization}, which shows boxplots of top-1 accuracy with different initialization types.
As expected, utilizing 
distinct pretrained models for the teacher and student network leads to better performance. 
Our analysis suggests that it is crucial to initialize the student and teacher network with different models to benefit from warm-starting the student model.

\subsection{Different Bayesian Optimization Methods}
We verify the effectiveness of \ours with respect to various BO methods. We perform additional experiments with Gaussian Process (GP) \cite{rasmussen2006gaussian} and Sequential Model-based Algorithm Configuration (SMAC) \cite{hutter2011sequential}. 
We utilize publicly available implementations \cite{smac3} of these methods and adopt their default settings. \autoref{table:various-bos} presents the results of our experiments, showing that all BO methods improve the performance of the baseline (i.e. 74.43). \ours further brings considerable performance improvement in all methods, demonstrating its effectiveness across different BO algorithms.

\begin{table}[t]
\centering
\begin{tabular}{lccc}
\toprule
 & TPE & GP & SMAC  \\
\midrule
BO & 76.48 & 75.69 & 75.93 \\ 
\ours (Ours) & \textbf{77.69} & \textbf{77.36} & \textbf{77.54} \\ 
\bottomrule
\end{tabular}
\vspace{6pt}
\caption{Comparison of different Bayesian optimization methods on CIFAR-100 with VGG-16.}
\label{table:various-bos}
\end{table}

\subsection{Various Distillation Techniques}
\begin{table}[t]
\centering
\begin{tabular}{lcccc}
\toprule
 & MSE & KL-div. & FitNets & Attention \\
\midrule
SD & 76.08 & 75.63 & 76.19 & 75.93 \\ 
\ours (Ours) & \textbf{77.69} & \textbf{77.32} & \textbf{77.81} & \textbf{77.35} \\ 
\bottomrule
\end{tabular}
\vspace{6pt}
\caption{Comparison of different distillation techniques on CIFAR-100 with VGG-16.}
\label{table:various-losses}
\end{table}
We finally evaluate the scalability of \ours by examining its performance with different distillation techniques. 
Specifically, we further investigate the efficacy of \ours with three popular distillation methods: Kullback-Leibler (KL) divergence \cite{hinton2015distilling}, FitNets \cite{romero2014fitnets}, and Attention \cite{zagoruyko2016paying}. 
We follow the suggestion in the original papers for the additional hyperparameters introduced by each distillation loss.
\autoref{table:various-losses} shows the results of the various distillation methods. We observe that SD successfully enhances the performance of the baseline (i.e. 74.43) across all tested distillation methods, and \ours further improves the performance significantly. These findings demonstrate that the benefit of \ours is not limited to specific distillation methods but can be extended to various techniques.

\section{Conclusion}
\label{sec:conclusion}


In this paper, we present \ours framework, which combines Bayesian Optimization (BO) and Self-Distillation (SD) to fully leverage the model, hyperparameter configuration, and performance knowledge acquired during the BO process, across all trials.
Through extensive experiments in various settings and tasks, we demonstrate that \ours achieves significant performance improvements, that is consistently better than standard BO or SD on their own.
\ours does not impose any additional overhead at test time and is versatile in that it is applicable to different \hs{BO and SD} methods.
Based on the presented evidence, we believe that the concept of marrying BO and SD is a powerful approach to training models, that should be further explored by the research community.

{\small
\bibliographystyle{ieee_fullname}
\bibliography{main}
}

\end{document}


\title{Bayesian Optimization Meets Self-Distillation (Supplementary Materials)}

\author{
HyunJae Lee\thanks{Authors contributed equally.} \quad Heon Song$^{*}$ \quad Hyeonsoo Lee$^{*}$ \quad Gi-hyeon Lee \quad Suyeong Park \quad Donggeun Yoo\\
Lunit Inc.\\
{\tt\small \{hjlee, hslee, dgyoo\}@lunit.io,}
{\tt\small \{songheony, lghsigma597, suyeong.park0\}@gmail.com}\\
}

\maketitle
\ificcvfinal\thispagestyle{empty}\fi

\appendix
\newcommand{\paperref}[1]{\textcolor{blue}{#1}}

\section{More Comparisons with Various Methods}
\begin{table}[t]
\centering
\addtolength{\tabcolsep}{-1pt}
\begin{tabular}{lccc}
\toprule
Method & CIFAR-10 & CIFAR-100 & Tiny-ImgNet \\
\midrule
Baseline & 93.75 & 74.43 & 53.33 \\
Grid & 93.87 & 74.51 & 53.94 \\
\midrule
SD & 94.21 & 76.08 & 56.46  \\
BO & 94.52 & 76.48 & 55.39  \\
\midrule
SD+BO & 94.57 & 76.53 & 56.89 \\ 
BOHB & 94.66 & 76.64 & 56.13  \\ 
\ours (Ours) & \textbf{94.98} & \textbf{77.69} & \textbf{58.55}  \\ 
\bottomrule
\end{tabular}
\vspace{2pt}
\caption{Extended comparison of Top-1 accuracy (\%) on CIFAR-10/100 and Tiny-ImageNet with VGG-16, incorporating additional methods like Grid, SD+BO, and BOHB.} 
\vspace{-1mm}
\label{table:more_methods}
\end{table}
Expanding upon the experiments conducted in \paperref{Section 4.1}, our comparative analysis is extended by incorporating additional methods on CIFAR-10/100 and Tiny ImageNet datasets, namely Grid, SD+BO, and BOHB. The Grid method conducts hyper-parameter searches within the same search space and budget as the other methods. The SD+BO employs the standard SD process but integrates BO directly into the training of the student model. Additionally, for benchmarking \ours against a state-of-the-art approach, we select BOHB~\cite{falkner2018bohb}, recognized for its exceptional performance across various benchmarks through the fusion of BO and Hyperband. The results of the comparative analysis are consolidated in \autoref{table:more_methods}. While both SD+BO and BOHB exhibit performance enhancements over individual SD and BO strategies, \ours demonstrates superior performance by effectively leveraging prior knowledge.

\section{Extra Computational Cost}
\ours involves multiple SD trials throughout the BO process. Despite the increased computational resource requirements, BO is widely adopted in numerous applications \paperref{[6,24,25]} due to its ability to enhance task performance while minimizing manual hyperparameter search. Following a similar principle, \ours proposes an effective design choice that combines BO and SD for a substantial boost in model performance. 
\tabcolsep=0.04cm
\begin{table}[t]
\centering
\begin{tabular}{llll}
\toprule
 & CIFAR10 & CIFAR100 & Tiny-ImgNet \\
\midrule
{\footnotesize SD+BO} & 1h 11m 2s & 1h 12m 38s & 3h 34m 40s \\ 
BOSS & 1h 11m 4s (+2s) & 1h 12m 41s (+3s) & 3h 34m 42s (+2s) \\ 
\bottomrule
\end{tabular}
\vspace{1pt}
\caption{Training duration of VGG-16 for a single trial on one T4 GPU, 4 core Intel Skylake CPU and 15GB RAM.}
\label{table:training_duration}
\vspace{-10pt}
\end{table}
While \ours introduces a negligible amount of additional training time compared to the naive combination of SD and BO (as shown in \autoref{table:training_duration}), it shows meaningful performance improvement over SD+BO (see \autoref{table:more_methods}).
More importantly, \ours introduces no further computational burden during inference. Therefore, when the deployed model is used in practical applications, there is no extra computational overhead.
Given its performance benefits, analogous to the established value of BO in diverse machine learning applications, \ours stands as a worthwhile extension despite the slightly higher computational demand.

{\small
\bibliographystyle{ieee_fullname}
\bibliography{main}
}